\documentclass[lettersize,journal]{IEEEtran}
\usepackage{amsmath,amsfonts}
\usepackage{algorithmic}
\usepackage{algorithm}
\usepackage{array}
\usepackage[caption=false,font=normalsize,labelfont=sf,textfont=sf]{subfig}
\usepackage{textcomp}
\usepackage{stfloats}
\usepackage{url}
\usepackage{verbatim}
\usepackage{graphicx}
\usepackage{cite}
\usepackage{comment}
\usepackage{multirow}
\usepackage{makecell}
\newcolumntype{P}[1]{>{\centering\arraybackslash}p{#1}}
\newcommand*{\MyIndent}{\hspace*{0.5cm}}

\begin{document}

\title{Learning Sparse Temporal Video Mapping for Action Quality Assessment in Floor Gymnastics}

\author{Sania Zahan, Ghulam Mubashar Hassan, ~\IEEEmembership{Member,~IEEE,} Ajmal Mian, ~\IEEEmembership{Senior Member,~IEEE}
\thanks{The authors are with the Department of Computer Science and Software Engineering, The University of Western Australia, Crawley WA 6009, Australia (e-mail: sania.zahan@research.uwa.edu.au (corresponding author); ghulam.hassan@uwa.edu.au; ajmal.mian@uwa.edu.au).}
}



\maketitle

\begin{abstract}
Athlete performance measurement in sports videos requires modeling long sequences since the entire spatio-temporal progression contributes dominantly to the performance. It is crucial to comprehend local discriminative spatial dependencies and global semantics for accurate evaluation. However, existing benchmark datasets mainly incorporate sports where the performance lasts only a few seconds. Consequently, state-of-the-art sports quality assessment methods specifically focus on spatial structure. Although they achieve high performance in short-term sports, they are unable to model prolonged video sequences and fail to achieve similar performance in long-term sports. To facilitate such analysis, we introduce a new dataset, coined AGF-Olympics, that incorporates artistic gymnastic floor routines. AFG-Olympics provides highly challenging scenarios with extensive background, viewpoint, and scale variations over an extended sample duration of up to 2 minutes. In addition, we propose a discriminative attention module to map the dense feature space into a sparse representation by disentangling complex associations. Extensive experiments indicate that our proposed module provides an effective way to embed long-range spatial and temporal correlation semantics.
\end{abstract}


\section{Introduction}
Action quality assessment (AQA) in sports videos aims to evaluate the quality of the performed action \cite{Pirsiavash_ECCV_2014, Yu_2021_ICCV, Tang_2020_CVPR}. AQA can be used for a number of sports to provide assessments with higher accuracy and consistency while reducing or even completely removing human involvement.  
For example, it can be employed to predict scores of sports events, rank athletes based on their moves, or analyze their moves to estimate future injury risks. Therefore, it can assist in a number of crucial factors in sports including self-assessment, performance analysis, player selection, and objective judgment. AQA provides affordable self-assessments for aspiring athletes to help them evaluate and improve their performance. 

Building a robust AQA method requires a diverse dataset encompassing elaborate scenarios and variations. However, existing AQA datasets have limitations in different aspects. Firstly, most datasets are collected from a small number of events. This limits subject variations, which is crucial to design generalized AQA methods. The difference in skills and techniques among athletes is usually very narrow, and with fewer participants, it becomes challenging to demonstrate detailed attributes. Less number of events result in limited scene variations. A realistic AQA benchmark should attain large subject, action, and scene variations.

Secondly, most datasets have very restrictive camera viewpoints. Thus, an athlete is visible from only one side. Furthermore, since the camera focus and distance to the athlete are constant, there are no scale variations. The third factor is that existing datasets only provide single-modality data. Thus existing AQA methods are unable to exploit modality-specific techniques or multimodal learning.

Finally, the most important limitation of existing datasets is the short duration of the videos. Samples in those datasets usually last a few seconds. Thus, designing advanced techniques to model long-term semantics is not possible. 

\begin{figure}[t!]
\centerline{\includegraphics[width=6cm]{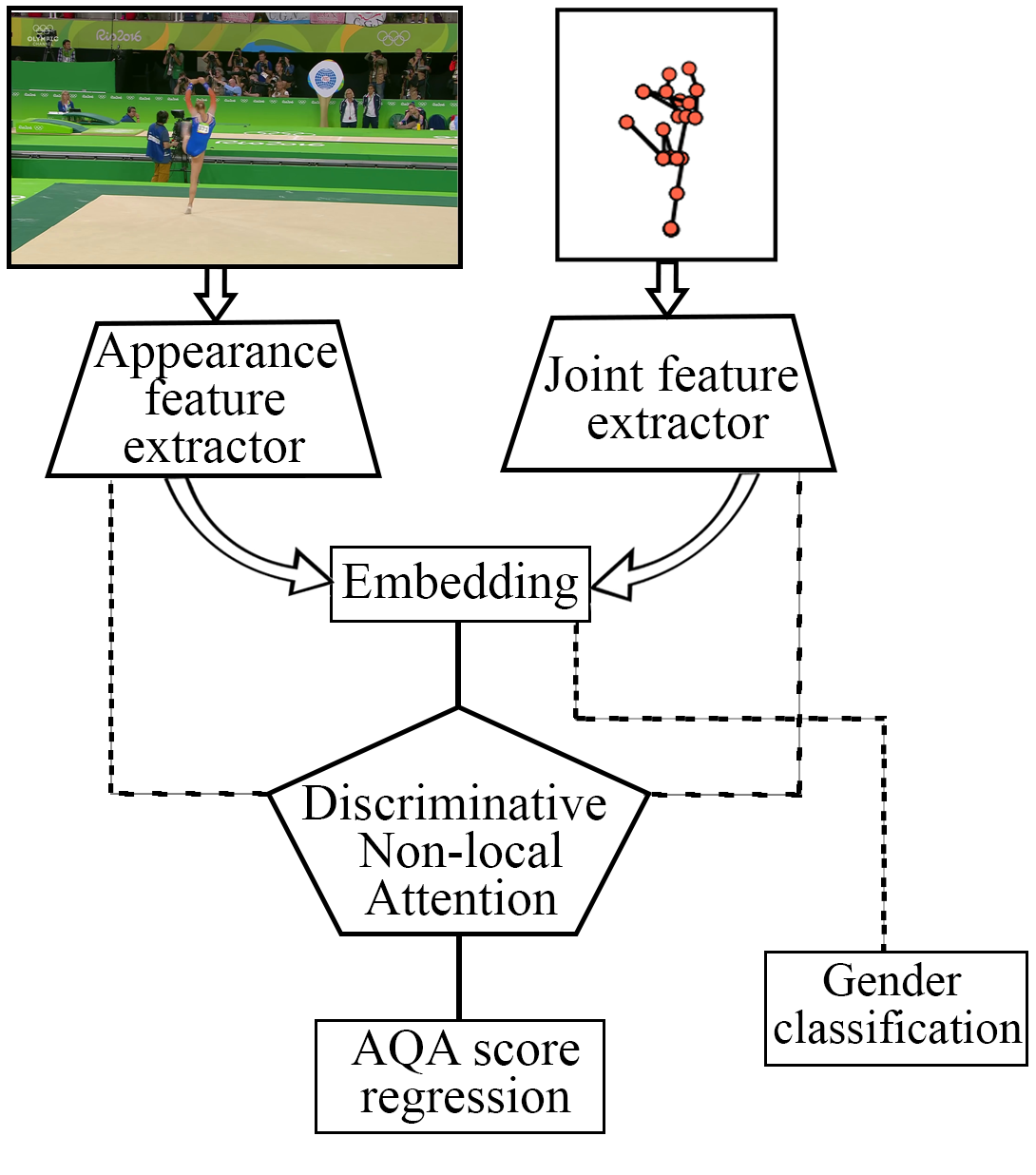}}
\caption{AQA conceptual workflow: discriminative non-local attention focuses on latent spatio-temporal association. The gender classification head forces the model to learn the distinctive features of female and male athletes.}
\label{work_flow}
\end{figure} 

\IEEEpubidadjcol

To overcome these limitations we propose a new dataset, AGF-Olympics. Our dataset comprises artistic gymnastic floor routines with around 1.5 minutes of video duration. Since floor routines demonstrate highly complex maneuvers, they can provide the best opportunity to study intricate body part movements and their association with action quality.

Ours is the only AQA dataset that provides the locations of athlete skeletal joints in each video sample. In gymnastic floor routines, an athlete performs many complex exercises where specific positions of the head, hands, and legs are critical to assess execution level. For example, the same double-twisting double-backs done by two athletes will achieve different scores depending on the rigidity of their bodies. Skeleton poses provide exact body positions and can be used to encode perceptual creativity representing subtle skill variations. However, visual information is also necessary. Videos give details of the floor and other visual cues, such as ``steps over the boundary line" penalties. Moreover, dynamic camera movement creates extensive view and scale variations. Our dataset is collected from the largest number of events, thus providing the highest subject variations as well. 

Our second contribution is a deep learning architecture for AQA. Existing human action-related methods \cite{MinsiW_2020_PAMI, liu2020disentangling, YuxinC_2021_ICCV, Dave_2022_CVPR} usually focus on classifying atomic action videos, performed with a few body part movements. A single frame is often enough to identify the actions in these simple videos. However, sports videos are more challenging as they incorporate several atomic actions with complex movements. Furthermore, the quality of these actions is spatially and temporally localized. Spatial distribution is often incomprehensible, and temporal characteristics dominate the overall quality. Therefore, these action classification methods are not optimal for sports AQA. 

Sports AQA focuses on analyzing the entire spatio-temporal associations. Analyzing comprehensive associations has some inherent challenges. For example, assessment in gymnastics will severely degrade due to a missed step or incorrect posture at any moment, resulting in a lower score. An AQA method must consider these incidents. Occlusion is another challenge with sports videos. Athletes' body parts are often occluded by nearby spectators, objects, or overlaid advertisements. Also, when the camera focuses on the athlete's face, it is hard to interpret what the athlete is doing in those frames. Therefore, an AQA method must automatically disregard these irrelevant frames while tracking performance measurement over the entire progression. Furthermore, sports videos have a wide range of complexity in motions and postures. It is essential to determine the local and global spatio-temporal relationships. AQA methods must connect these local individual patterns with global trends to evaluate fine-grained performances.  	

Existing AQA methods can be divided into two research directions: injury and score prediction. For injury prediction, researchers primarily focused on analyzing irregular reaction forces and moments to assess non-contact injury risks  \cite{Johnson_2018_MBEC, Johnson_2019_TBE, Johnson_2019_EJB, Johnson_2021_TBE, Mundt_2021_sensors, Goldacre_2021_ISBSC, morris_2021_ISBSC}. For score prediction,  end-to-end deep learning models are usually employed to map sports videos to final scores \cite{Parmar_2017_CVPR_Workshops, Bertasius_2017_ICCV, Doughty_CVPR_2018, Yongjun_2018, Parmar_2019_CVPR, Parmar_2019_WACV,  Xu_TCSVT_2020, Tang_2020_CVPR, Yu_2021_ICCV}. Some researchers introduced the Gaussian distribution of score to incorporate uncertainty \cite{Tang_2020_CVPR} or look-up tables to leverage predefined rules \cite{Nekoui_2020_CVPR_Workshops}. These techniques provide additional information in the learning procedure. 

However, these AQA methods overlook the importance of mapping longer temporal semantics and only focus on short-duration sports, where videos are mostly captured from a single viewpoint. Long-duration sports have an extended temporal range, and mapping extended semantic correlation is a critical limitation in current AQA methods. For example, videos of gymnastic floor routines have dynamic viewing angles captured using multiple cameras. These videos usually extend to a couple of minutes, representing highly complex postures compared to sports events like diving and gymnastic vault \cite{Yongjun_2018, Parmar_2019_WACV, pan_iccv_2019, Parmar_2019_CVPR, Nekoui_2020_CVPR_Workshops, Tang_2020_CVPR, Yu_2021_ICCV}.

To address this particular aspect, we propose a deep learning module: Discriminative Non-Local Attention (DNLA), that focuses on learning long-term semantics. DNLA extracts the most prominent features localized in spatio-temporal space and eliminates less informative features. Thus it concentrates the learning exclusively on identifying intricate relationships that cause the highest impact on the final score. 

Figure~\ref{work_flow} illustrates our overall workflow at the conceptual level. In addition to score regression, our gender classification module enhances focus on gender-specific subtleties. Floor routines have some differences based on gender. Female athletes must combine artistry with the presentation of force and techniques, while male athletes only focus on the latter. Thus gender classification enables the model to comprehend inherent distinctions. 

We also present a conceptual design of an AQA app. This app can be used by athletes to measure progress and compare techniques. Since our dataset incorporates only Olympic video samples, the AQA model deployed in the app is capable of providing Olympic-level score predictions. 

Our main contributions are summarized as follows:
\begin{itemize}
  \item \textbf{New dataset:} We propose a new dataset AGF-Olympics that contains videos and athlete skeletal locations in each frame along with their scores. We collected video samples from Olympic events that reflect difficulty and execution at their best. Our dataset provides challenging real word scenarios with diverse camera views, subjects, and occlusions. The skeleton sequences in our dataset allow researchers to analyze intricate structural pose variations.
  
  \item \textbf{Discriminative Non-local Attention module:} Our proposed learning module creates a sparse temporal mapping to enhance semantic correlation over an extended temporal range for AQA in sports videos. It enables extracting distinguished global semantic associations which are often subdued in long-range temporal progression.   
  
  \item \textbf{Experimental results:} Experimental results on the AGF-Olympics dataset with existing methods indicate clear challenges. Moreover, comparative evaluations of our proposed method demonstrate the advantages of selective sparse attention. 
\end{itemize}

The remaining paper is organized as follows: Section~\ref{litreature_rev} describes existing AQA methods. Section~\ref{Dataset} represents the proposed dataset. Section~\ref{Methodology} illustrates details of the proposed architecture. Section~\ref{Experiments} provides experimental results, comparisons with existing methods, and ablation analysis. Conclusions are provided in Section~\ref{Conclusions}.

\section{Related Work}\label{litreature_rev}
In this section, we explore publicly available AQA datasets and existing methods.

\subsection{Action Quality Assessment datasets}
MIT-Dive and MIT-Skate \cite{Pirsiavash_ECCV_2014} are one of the first AQA datasets. MIT-Dive has 159 samples. All videos in the dataset are captured at a 60 fps frame rate with 150 average number of frames per sample. The final quality scores range between 20 to 100. Whereas, MIT-Skate contains 150 sample videos. The frame rate is 24 fps, and the quality scores range between 0 to 100. Another dataset called UNLV-Dive \cite{Parmar_2017_CVPR_Workshops} is an extended version of MIT-Dive. It contains an additional 211 samples and also includes 176 vault samples named as UNLV-Vault. Many methods have been evaluated on these datasets, especially MIT-Dive and UNLV-Dive \cite{Pirsiavash_ECCV_2014, Parmar_2017_CVPR_Workshops, Parmar_2019_WACV, Xu_TCSVT_2020, Nekoui_2021_WACV}.

AQA-7 \cite{Parmar_2019_WACV} contains 7 types of sports: diving, vault, skiing, snowboarding, synchronized diving (3m and 10m), and trampoline. Diving is incorporated from the UNLV-Dive dataset. These sports are of a short sequence with a maximum of 25 seconds duration. 

MTL-AQA \cite{Parmar_2019_CVPR} includes diving videos. Along with action quality scores, this dataset provides commentaries and frame-level pose annotations. MTL-AQA has been used for multi-task learning. 

Fis-V is a figure skating dataset proposed in \cite{Xu_TCSVT_2020}. It provides two different scores, technical element (TES) and program component (PCS). Samples are collected only from ladies' single short program.


Although these datasets provide an opportunity to analyze action quality assessment methods, they have limitations in many aspects. For example, these datasets primarily focus on the diving action/sport e.g. MIT-Dive \cite{Pirsiavash_ECCV_2014}, UNLV-Dive \cite{Parmar_2017_CVPR_Workshops}, MTL-AQA \cite{Parmar_2019_CVPR} etc. Given the nature of the diving action, samples in these datasets are of very short duration i.e., about a few seconds. Also, diving videos are captured from a single-camera viewpoint. Thus these datasets are limited in terms of view and scale variations. Although MIT-Skate \cite{Pirsiavash_ECCV_2014} and Fis-V \cite{Xu_TCSVT_2020} contain long-duration sports, they are collected from a limited number of events, resulting in limited subject variations.   

To overcome the limitations of the existing dataset, we propose the AGF-Olympics which introduces some unique challenges. Our dataset incorporates long sequence videos with an average duration of 1.5 minutes duration (maximum 7051 frames), and significant variations in camera viewpoints, subjects, scales, and background scenes. Our dataset also provides skeletal joint locations in the video frames. Furthermore, we have collected samples from the longest period i.e., four Olympics and trials from 2008 to 2021. Table \ref{dataset_comparison} illustrates a comparison of existing datasets with AGF-Olympics.

\subsection{Video-based AQA methods}
Most AQA methods use raw video frames to predict scores. The C3D-LSTM method in \cite{Parmar_2017_CVPR_Workshops} performs clip-based feature aggregation with score prediction to evaluate the UNLV-Dive dataset. 
C3D-AVG-MTL method utilizes a multi-task approach for AQA score regression, caption generation, and factorized action classification on the MTL-AQA dataset \cite{Parmar_2019_CVPR}. However, the frame-level annotation of body poses for action classification and commentary scripts can limit the possibility of extension and transfer learning. The C3D-LSTM method in \cite{Parmar_2019_WACV}, utilizes clip-based C3D features with LSTM to learn temporal modeling. It was evaluated on six out of the seven different sports events in AQA-7. The trampoline was excluded due to its much longer duration. This highlights the difficulty of modeling sports actions over an extended temporal range.

USDL method uses a multi-pathway to predict multiple score distributions provided in AQA-7, emulating scoring systems in sports by multiple judges \cite{Tang_2020_CVPR}. CoRe + GART exploits contrastive learning to discriminate subtle differences in samples \cite{Yu_2021_ICCV}. It is evaluated on AQA-7 and MTL-AQA. 
C3D + S-LSTM + M-LSTM illustrated using the multi-scale convolutional skip LSTM to capture sequential information in the long-range figure skating videos in Fis-V \cite{Xu_TCSVT_2020}. However, it reports a high mean squared error (MSE) rate of 19.91, indicating a huge difference from actual scores. 

\subsection{Pose based AQA methods}
Few AQA methods exploit the distinct characteristics of pose features. The Pose + DCT method applies hierarchical independent subspace analysis and DCT/DFT features from poses to assess quality \cite{Pirsiavash_ECCV_2014}. 
The joint difference and joint commonality modules in \cite{pan_iccv_2019} use patch videos of major body joints from AQA-7 to extract motion features to represent better joint dependency. Nonetheless, cropping local patches around joints is very time-consuming. 
The FALCONS method incorporates a difficulty assessor that uses a look-up table to assess the difficulty level of sub-activities performed during diving \cite{Nekoui_2020_CVPR_Workshops}. EAGLE-Eye, on the other hand, uses a joint coordination assessor on pose heatmaps to learn temporal dependencies using multiscale convolution \cite{Nekoui_2021_WACV}. 

Experimental results indicate that these methods have insufficient spatial and temporal coordination over an extended time to achieve similar performance for longer sports. Therefore, we address these issues in this work by focusing on extreme-level pose contortion and long video sequences. 

\section{Proposed AGF-Olympic AQA Dataset}\label{Dataset}
In this section, we describe our proposed AGF-Olympic 
dataset. We introduce detailed data collection strategies and related statistics. 

\subsection{Data Collection:} We started the data collection procedure by surveying broadcast videos of gymnastic floors on different video-sharing websites. We carefully skimmed through hundreds of videos to ensure that only high-standard videos are selected that contain skilled performances. For example, amateur-level gymnastic routines do not incorporate many highly valued skills, and the difficulty level is also low. On the other hand, professionals train for years before competing in the Olympic games. Therefore, we selected events only from the Olympic trials, qualifying rounds, and finals to acquire high-quality samples with rich skill variations. We also ensured data collection with a range of action quality. 

Since 2008, athletes are scored in three areas: difficulty, execution, and penalty instead of the legacy scoring on a scale of 10. Therefore, we did not include events before 2008 to keep uniformity in the scoring system. The dataset comprises all gymnastic floor routines from the individual, all-around, teams, qualifications, and trial events in both women's and men's categories. We discarded samples where athletes could not complete a routine due to injuries or other interruptions. 

\subsection{Data Modality:} 
\begin{figure}[!htbp]
\centerline{\includegraphics[height=1.6in, width=3in]{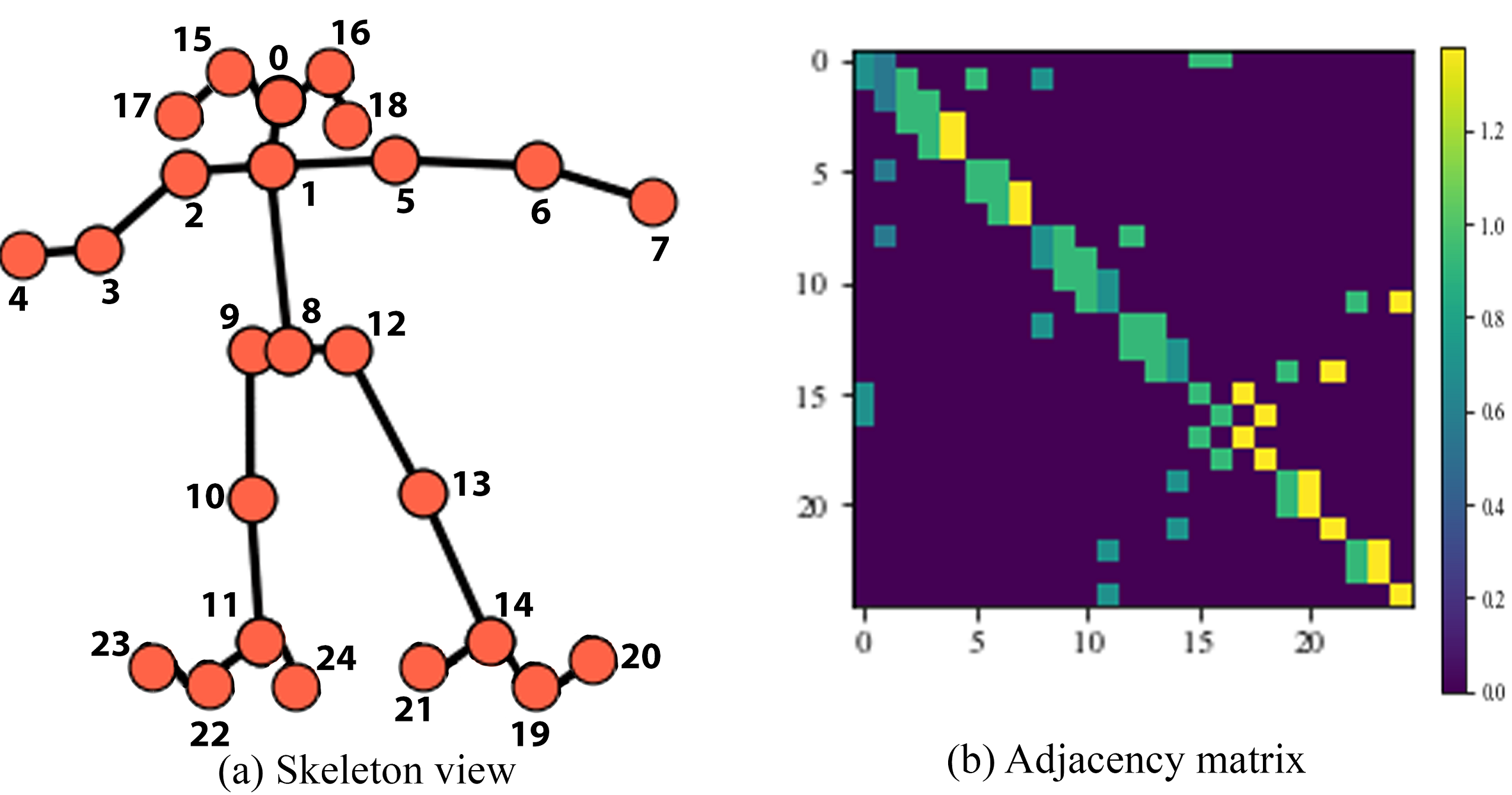}}
\caption{Configuration of 25 body joints extracted from videos using OpenPose \cite{openpose2019}. (a) Skeleton view with corresponding joint indexes (b) Adjacency matrix representing the joint neighborhood.}
\label{skeleton_OP}
\end{figure}

Our dataset provides two modalities: RGB videos and 2D joint position information (skeleton representation). We meticulously trimmed all samples at the start and end of each complete routine and used FFmpeg to extract individual clips. Then we used OpenPose to detect 2D skeleton joints of the athletes in videos \cite{openpose2019}. 
We discarded the frames that did not contain any skeleton. We stored the video files in mp4 and the skeletons in JSON format. Figure~\ref{skeleton_OP} (a) demonstrates the configuration of the body joints, and (b) illustrates the joint adjacency matrix.

\begin{figure*}[!htbp]
\centerline{\includegraphics[width=\linewidth]{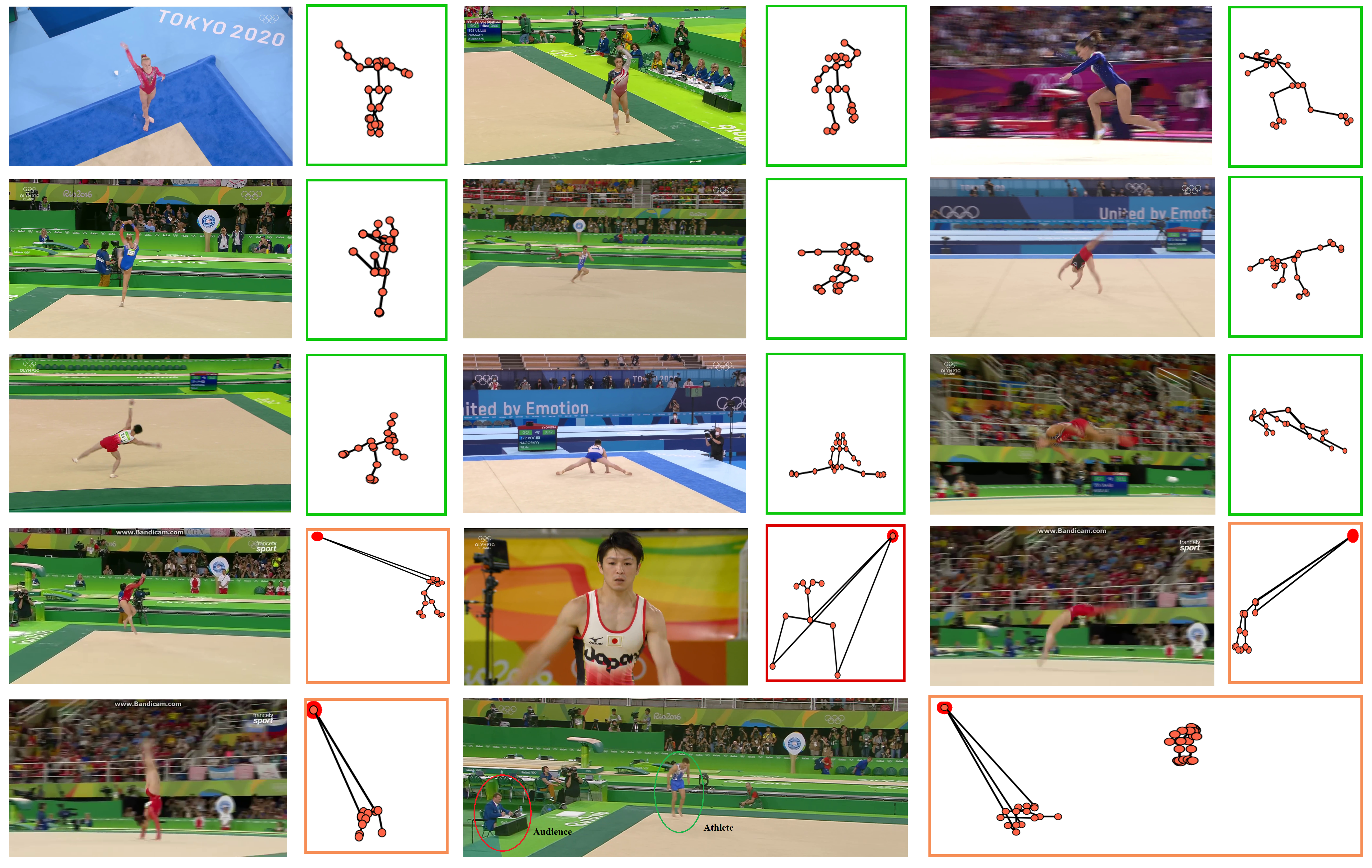}}
\caption{Sample video frames and corresponding skeleton frames of the AGF-Olympic dataset. The first three rows show a variety of subjects and camera views. The fourth row illustrates (from left) missing body joints (red circle at the origin (0,0) in skeleton view), scale variation due to upper body focus causing more than 50\% joints missing, and huge scale differences in the detected joints compared to others. The last pair in the fourth row and the first in the fifth row represent blurry samples. The final pair in the fifth row demonstrates incorrect skeleton detection. The skeleton of an audience gets incorrectly detected as the athlete due to having higher confidence. Green boxes indicate that the skeletons are accepted without any change, orange boxes indicate that the missing joints have been interpolated and red boxes indicate frames have been discarded.}
\label{samles_from_dataset}
\end{figure*}

Figure~\ref{samles_from_dataset} illustrates samples from our dataset in both RGB frames and corresponding skeleton poses. The first three rows represent diverse poses of female and male athletes from different camera viewpoints. The last two rows demonstrate four types of error frames, missing body joints (due to self-occlusion), massive scale difference, blurriness (due to rapid movements), and erroneous skeleton detection (detecting joints of an audience with higher confidence than the athlete).

\subsection{Data diversity:} Compared to most existing AQA datasets, which focus on a monotonic camera view, our dataset provides a challenging scenario with a highly uncertain viewpoint. Since we collected samples from broadcast videos, they are a composition of clips captured from different cameras with high movements to keep the athletes on a clear vantage point. Thus, samples contain higher disparity in viewpoints such as top, left-side, right-side, back, or just upper body view. 

Some camera viewpoints focus exclusively on the upper body of an athlete. Thus, video frames with these viewpoints fail to generate a meaningful skeleton. We discarded these skeletons along with the corresponding RGB video frames. Besides, cameras often focus solely on the athlete, causing that person to cover most parts of the video frames. A broader viewing angle, on the other hand, causes an athlete to cover only a tiny fraction of the video frames. It causes highly random scale variations due to continuous changes in body/frame ratio. Moreover, our dataset contains diverse and complex poses such as twists, flips, and somersaults. Existing AQA datasets that incorporate diving or vault do not have these challenges.

\subsection{Data Statistics:} We downloaded a total of 83 hours of videos. Since original broadcasting videos have varied resolutions up to 1920x1080 pixels, we restricted all samples to have 256x256 pixels resolution. Video duration is around 1.3 to 2.0 minutes with an average of 2500 frames. The dataset includes three types of annotations: difficulty, execution, and penalty scores, except for the all-around event, as they only provide total scores. Total scores have a standard deviation of 0.88. In case scores were missing from broadcast videos, we cross-checked Olympic official websites and other online resources to obtain the correct scores. Each sample comes with three additional pieces of information: event, gender, and year. These can be used to generate different evaluation criteria or learning techniques.

\begin{figure}[!hbtp]
\centerline{\includegraphics[width=\columnwidth]{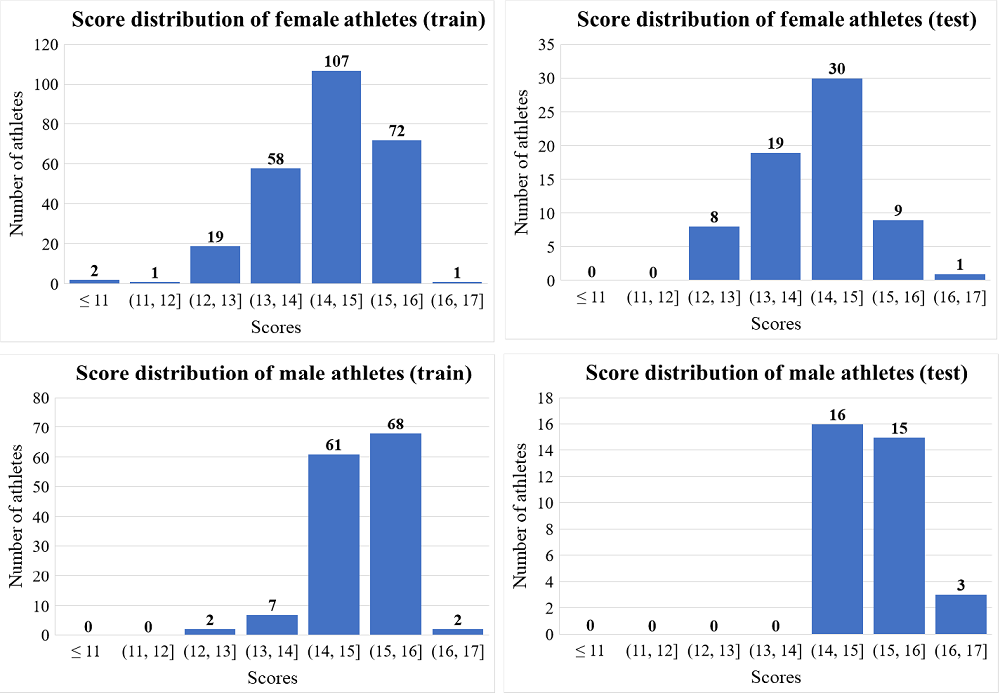}}
\caption{Score distribution in our dataset: top row train (left) and test (right) splits of female athletes and bottom row train (left), and test (right) splits of male athletes.} 
\label{score_distribution}
\end{figure}

Figure~\ref{score_distribution} shows the distribution of final scores in train and test splits for female and male athletes. The total number of samples for the floor routine is 500: female athletes 326, and male athletes are 174.

Table \ref{dataset_comparison} compares our proposed dataset with the existing AQA dataset. Our dataset covers the highest number of events spanning four Olympics, and with a challenging intra-sample scale and view variations with two different data modalities.

\begin{table*}[!hbtp]
\scriptsize 
\centering
\caption{Comparison of our proposed and existing AQA datasets\label{dataset_comparison}}
\begin{tabular}{|p{55pt}|cc|P{42pt}|P{30pt}|P{25pt}|ccc|P{27pt}|P{36pt}|P{55pt}|}
\hline
\multirow{2}{*}{\textbf{Dataset}} & \multicolumn{2}{c|}{\textbf{Modality}}       & \multirow{2}{*}{\textbf{Gender}} & \multirow{2}{*}{\textbf{Samples}} & \multirow{2}{*}{\makecell{\textbf{Events}}} & \multicolumn{3}{c|}{\textbf{Variations}}  & \multirow{2}{*}{\makecell{\textbf{Year} \\ \textbf{range}}} & \multirow{2}{*}{\makecell{\textbf{No of frames}\\ \textbf{(max)}}} & \multirow{2}{*}{\textbf{Labels}} \\ \cline{2-3} \cline{7-9}
                         & \multicolumn{1}{c}{\textbf{RGB}} & \textbf{SKEL} &                         &                          &                         &    \multicolumn{1}{c}{\textbf{Scene}} & \textbf{View} & \textbf{Scale}                                                                                                                                                                                         &                            &                                                                             &                         \\ \hline
MIT Dive \cite{Pirsiavash_ECCV_2014}    & \checkmark  & $\times$  & Male  &  159 & 1 & $\times$ & $\times$ & $\times$ & 2012 & 151 & AQA score\\ \hline
 MIT Skate \cite{Pirsiavash_ECCV_2014}    & \checkmark  & $\times$  & Male, Female  &  171 & 13 & \checkmark & \checkmark & \checkmark & 2012 & 5823 & AQA score\\ \hline
UNLV-Dive \cite{Parmar_2017_CVPR_Workshops}    & \checkmark  & $\times$  & Male  &  370 & 1 & $\times$ & $\times$ & $\times$ & 2012 & 151 & AQA score\\ \hline
UNLV-Vault \cite{Parmar_2017_CVPR_Workshops}    & \checkmark  & $\times$  & Male, Female  &  176 & 7 & $\times$ & $\times$ & $\times$ & 2012-15 & 100 & AQA score\\ \hline
AQA-7 \cite{Parmar_2019_WACV} & \checkmark  & $\times$  & Male  &  1189 & - & \checkmark  & $\times$ & $\times$ & - & 618 & AQA score\\ \hline
MLT AQA \cite{Parmar_2019_CVPR} & \checkmark  &  $\times$  & Male, Female &  1412 & 16 & \checkmark & $\times$ & $\times$ & 2012-14 & 274 & AQA score, Action Class, Caption\\ \hline
Fis-V \cite{Xu_TCSVT_2020} & \checkmark  &  $\times$  & Female &  500 & - & \checkmark & \checkmark & \checkmark & - & 4300 & AQA score\\ \hline
\textbf{AGF-Olympics (Ours)}    & \textbf{\checkmark} & \textbf{\checkmark}  & \textbf{Male, Female}  & \textbf{500} & \textbf{35} & \textbf{\checkmark} & \textbf{\checkmark} & \textbf{\checkmark} & \textbf{2008-20} & \textbf{\makecell{7051}} & \textbf{AQA score, Gender, Difficulty} \\ \hline   
\end{tabular}
\end{table*}

\section{Proposed Architecture}\label{Methodology}
\begin{figure*}[!htbp]
\centerline{\includegraphics[width=\linewidth]{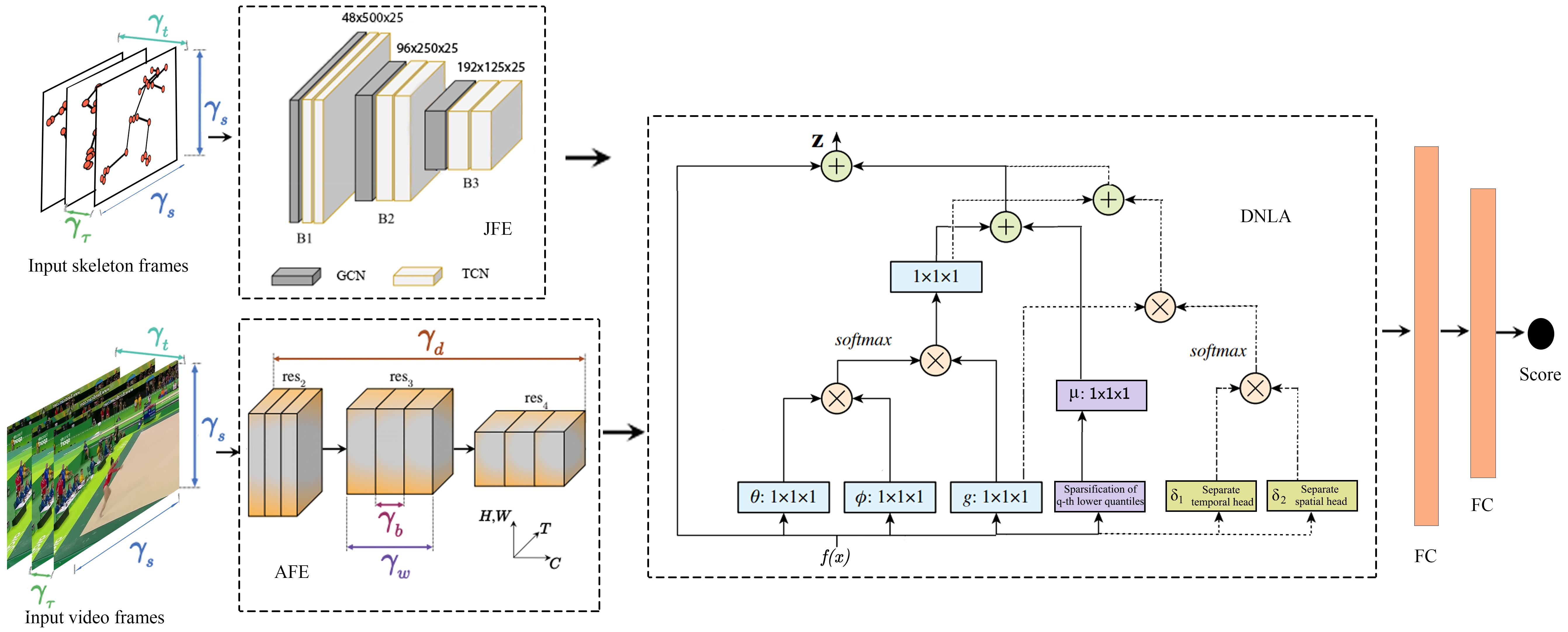}}
\caption{Overview of our architecture: Input frames are segmented into 7 non-overlapping clips of $T$ frames. JFE extracts features from skeleton joint clips, and AFE extracts appearance features from RGB clips. DNLA models subtle long-term spatio-temporal dependency and creates a sparse feature representation from the dense feature embedding coming from the base encoder (JFE or AFE). Finally, MLP maps feature space to the final score.}
\label{proposed_architecture}
\end{figure*}

In this section, we introduce our proposed AQA model which includes three modules: feature encoder that embeds raw frames to a higher dimensional matrix subspace, DNLA that generates a sparse representation, and MLP that creates the final mapping. Figure~\ref{proposed_architecture} illustrates the overall architecture. The following sections describe the technical functionalities of each module.

\subsection{Joint Feature Encoder (JFE)}
We experimented with two GCN architectures as JFE, a simple GCN model and MSG3D \cite{liu2020disentangling}, to extract detailed joint interactions. The first GCN model includes one spatial graph convolution layer followed by two separable temporal convolutions. It uses a physical joint adjacency matrix with layer-wise scalar dynamic adaptation. MSG3D is a more advanced GCN model using multi-scale aggregation with unified spatio-temporal modeling to extract complex relationships. Joint adjacency is critical in understanding activities, and scaled \emph{k}-adjacency matrix in MSG3D allows a more efficient representation. We used features from the last GCN layers of JFE as input to our DNLA module. 

\subsection{Appearance Feature Encoder (AFE)}
We experimented with X3D \cite{x3d_2020} and I3D \cite{Carreira_2017_CVPR} networks as our appearance feature encoder module. The progressive expansion of x3D with lower channel dimensions and higher spatio-temporal resolution allows effective video feature extraction. We extracted features from the last convolutional layer (conv5) of X3D. A similar process has been used in the case of I3D. Features for each clip are combined along the temporal dimension to generate the final feature matrix.

\subsection{Discriminative Non-local Attention (DNLA)}
Long-range dependencies are often captured using large receptive fields in convolutional or recurrent operations. Repetitive local calculations are inefficient, and optimization can be arduous. It becomes significantly complex if the relationship between distant positions in space and time needs to be addressed. Classical non-local mean operation \cite{Buades2005} assumes that under stationary conditions, patches in an image have a certain similarity, and the similarity details increase as the image size grows. Thus a pixel can be estimated from neighboring pixels in the image. This phenomenon extends to videos as well. Particularly, sports videos have several overlapping moments with similar human activities. Non-local operations can be exploited to capture these distant dependencies. It is orthogonal to self-attention in transformer \cite{NIPS2017_transformer} that has achieved tremendous success in machine translation. 

Non-local network \cite{Non_local_wang} uses non-local mean to calculate the response of a spatio-temporal pixel position as a weighted sum of all pixels over the spatial and temporal dimensions. Thus the calculated features capture relation over the entire space-time instead of a local patch provided by kernels in convolution or consecutive frames in recurrent networks.

Gymnastic floor routines usually have a large number of frames. Traditional video or skeleton-based architectures are not designed to handle such extended temporal modeling. These architectures usually evaluate samples with very short lengths, lasting a few seconds. To process longer videos, existing works use independent segments of a video and concatenate the segment-wise encoded features. It results in an asynchronous dense feature matrix. This method works if the video does not contain complex spatio-temporal dependency or the downstream task is simple, such as action recognition. But it fails to demonstrate equal performance for more elaborate tasks such as action quality assessment. To address the discrepancy in the segments, we introduce discriminative non-local attention that computes correlation among distant feature positions. Thus it creates a sparse mapping by extracting the most prominent features over the temporal dimension. 

Our vanilla feature distillation module applies an adaptive average pooling with an overlapping stride and max pooling over the space-time feature space. Finally, the pooled channels are concatenated and passed through MLP for score mapping. We address it as VFD. Next, we use a non-local layer to enhance correlation significance among distant positions. A generic non-local calculation from \cite{Non_local_wang} is defined as in Eq~\ref{NL_block}.

\begin{equation}\label{NL_block}
 y_i = \frac{1}{C(x)}\sum_{\forall_j}f(x_i,x_j)g(x_j)
\end{equation}
Here, $i$ is the index of the output position, and $j$ represents all positions over space and time for the input $x$. The pairwise function $f$ performs scalar multiplication between $i$ and all $j$ positions. The unary function $g$ computes output at position $j$. The final output is normalized by a factor of $C(x)$. We evaluated two different pairwise functions $f$: Embedded Gaussian and Concatenation \cite{Non_local_wang}. Embedded Gaussian performed superior in our experiments. It is defined as in Eq~\ref{embd_G}. 

\begin{equation}\label{embd_G}
 f(x_i,x_j) = e^{{\theta (x_i)}^T{\phi(x_j)}},
\end{equation}
where, $\theta (x_i) = W_\theta x_i$ and $\phi (x_j) = W_\phi x_j$ are two embedded representation of the input matrix.

Our vanilla distillation module suppresses redundant or less informative attributes without losing distinguishing semantic details. On the other hand, non-local convolutional operation facilitates learning correlation among distant frames. We bridge these two learning procedures by unifying the idea of discriminative distillation of VFD into non-local convolution. We tested two different approaches: DNLA$_\mu$ and DNLA$_\delta$.

In DNLA$_\mu$, we combine a parallel branch with the original NL block that acts as an embedded residual connection. It takes the $g(x)$ of Eq~\ref{NL_block} as input and creates a motion matrix by subtracting each consecutive frame. Each frame in the new matrix expresses the joint motions and will have higher values whenever the athlete does an intense activity. We use a binary mask to deactivate the lower quantile of the matrix. It subdues irrelevant details and adds a dominant feature pipeline. After processing through an encoding layer, the embedded feature is fused into the main branch.  



The $\theta$ and $\phi$ operations in the non-local network (Eq~\ref{embd_G}.) extend the space-time domain into a generic embedding. DNLA$_\delta$ distributes the spatial and temporal feature spaces into two separate encoding heads. This independent computation incentivizes flexibility by allowing robust cross-spacetime interaction and augments non-local attention by retaining discriminative spatial and temporal dependency.

\section{Experiments}\label{Experiments}

We evaluate our proposed AQA method using our AGF-Olympics dataset. First, we discuss data preprocessing techniques and then the experimental results.

\subsection{Data preprocessing}
Skeletons extracted from sports videos are very noisy due to extreme movements. OpenPose \cite{openpose2019} often fails to detect skeletons in frames where the athlete is in high motion and appears blurry. Besides, sports replays often focus on the upper body of athletes, resulting in missing almost 50\% of the joint coordinates. We discard frames that have less than 20 joint coordinates. In frames where the number of missing joints is a maximum of 5, we used joint interpolation to calculate the missing joints from immediate neighbors instead of completely discarding the frames. We use K-hop association to identify neighboring joints in case immediate neighbors are also missing. Eq~\ref{joint_intp_eq} represents K-hop joint interpolation where $K$ is a set of neighboring joints, f is a function that chooses the closest two neighboring joints from the set K and i and j represent their indexes. 

\begin{equation}\label{joint_intp_eq}
X_{new} = \frac{f(K_i) + f(K_j)}{2};\qquad f(K_{i/j})\neq0\
\end{equation}


\subsection{Results}
We compare our method to existing AQA methods that achieved good  performance in predicting scores on the six sports of the AQA-7 dataset. These sports include diving, vault, skiing, snowboarding, synchronized diving 3m, and synchronized diving 10m. All these sports are of short duration, containing a maximum of 151 frames, and do not involve extended sequences of complex movements. These methods 
achieved on average 0.74 Sp. Corr. on the AQA-7 dataset with a minimum value of 0.60 and maximum of 0.88 whereas on our dataset average Sp. Corr. is 0.44 with a minimum value of 0.19 and a maximum of 0.61. It is clearly evident that when experimented on lengthy and complex sports, their performance is not on par. Details are represented in Table \ref{compare_methods_on_our_dataset}. The table also shows that our proposed method outperforms existing methods by a large margin and has a significantly smaller number of parameters.

\begin{table}[!htbp]
\scriptsize 
\caption{Performance comparison of different models on our proposed AGF-Olympics dataset.\label{compare_methods_on_our_dataset}}
\setlength{\tabcolsep}{3pt}
\centering
\begin{tabular}{|p{85pt}|P{50pt}|P{35pt}|P{43pt}|}
\hline
\textbf{Model} & \textbf{Architecture}	& \textbf{Model parameters}	& \textbf{Corr. on AGF-Olympic}\\
 \hline
C3D-LSTM \cite{Parmar_2017_CVPR_Workshops} & Full model & 65.67 M & 0.43 \\
 \hline
C3D-LSTM \cite{Parmar_2019_WACV} & Full model & 65.67 M & 0.43 \\
\hline
CoRe \cite{Yu_2021_ICCV} & Full model & 14.80 M & 0.19 \\
\hline
 Eagle-eye \cite{Nekoui_2021_WACV} & W/o JCA Stream & 84.22 M & 0.54 \\
\hline
USDL \cite{Tang_2020_CVPR} & Full model & 12.59 M & 0.61 \\
\hline
 \textbf{Proposed model} & \textbf{Full model} & \textbf{05.01 M} & \textbf{0.68} \\
\hline
\end{tabular}
\end{table}

Table \ref{JFE_ablate_results} reports experimental results with different architectural choices 
of our method 
on the skeleton sequences of AGF-Olympics. We conducted comprehensive ablation studies to evaluate the effectiveness of different architectural modules. For the base joint feature encoder, we used two different GCN models. First, we used a simple GCN model with no bells and whistles. Then we used a more advanced MSG3D model that achieved outstanding performance in the action recognition task \cite{liu2020disentangling}. MSG3D uses multi-scale and multi-window techniques with an inflated adjacency matrix. With everything else kept the same, MSG3D achieved a 14\% increase in Spearman correlation over the simple GCN model. 

Subsequently, we investigated different distillation processes to smooth the asynchronicity in the segmented feature and create a sparse mapping. Our vanilla distillation module VFD uses pooling techniques. Then we experimented with the non-local module that stimulates self-attention and enhances global modeling. Finally, we introduce discriminative distillation in the non-local module to intensify attention on the most contributing attributes. With the discriminative non-local attention module, MSG3D + DNLA$_\delta$(emb) achieved around 38.7\% increase in Spearman correlation compared to VFD.

\begin{table}[!htbp]
\scriptsize 
\caption{Results for different architectural choices in the joint stream. Base joint feature encoder (JFE): a GCN or MSG3D. VFD indicates the vanilla feature distillation and DNLA$_{\mu/\delta}$ indicates our proposed discriminative non-local attention module.\label{JFE_ablate_results}}
\setlength{\tabcolsep}{3pt}
\centering
\begin{tabular}{|p{88pt}|P{28pt}|P{20pt}|P{20pt}|P{18pt}|P{38pt}|}
\hline
\textbf{Method} & \textbf{Base Encoder} & \textbf{SP Corr} & \textbf{Test loss} & \textbf{\begin{tabular}[c]{@{}c@{}}Class\\ loss\end{tabular}} & \textbf{\begin{tabular}[c]{@{}c@{}}Joint\\ interpolation\end{tabular}}\\
\hline
 JFE + VFD + MLP & GCN & 0.27 & 1.36 & $\times$ & \checkmark\\
  \MyIndent + VFD + MLP & - & 0.32 & 1.33 & \checkmark & $\times$\\
 \MyIndent + VFD + MLP & - & \textbf{0.40} & 1.11 & \checkmark & \checkmark\\
\hline
 \MyIndent + VFD + MLP & MSG3D & 0.31 & 0.45 & $\times$ & \checkmark\\
 \MyIndent + NLA$_{(emb)}$ + MLP & - & 0.32 & 36.86 & $\times$ & \checkmark \\
 \MyIndent + NLA$_{(cat)}$ + MLP & - & 0.35 & 0.08 & $\times$ & \checkmark \\
 \MyIndent + DNLA$_{\mu_{(emb)}}$ + MLP & - & 0.34 & 0.04 & $\times$ & \checkmark \\    
 \MyIndent + DNLA$_{\mu_{(cat)}}$ + MLP & - & 0.35 & 0.08 & $\times$ & \checkmark \\    
 \MyIndent + DNLA$_{\mu_{(emb)}}$ + MLP & - &  0.41 & 0.04 & \checkmark & \checkmark \\    
 \MyIndent + DNLA$_{\delta_{(emb)}}$ + MLP  & - & 0.43 & 1.16 & $\times$ & \checkmark \\
 \MyIndent + DNLA$_{\delta_{(emb)}}$ + MLP  & - & \textbf{0.45} & 0.36 & \checkmark & \checkmark
 \\    
\hline
\end{tabular}
\end{table}

\begin{figure}[!hbtp]
\centerline{\includegraphics[height=1.5in, width=2.5in]{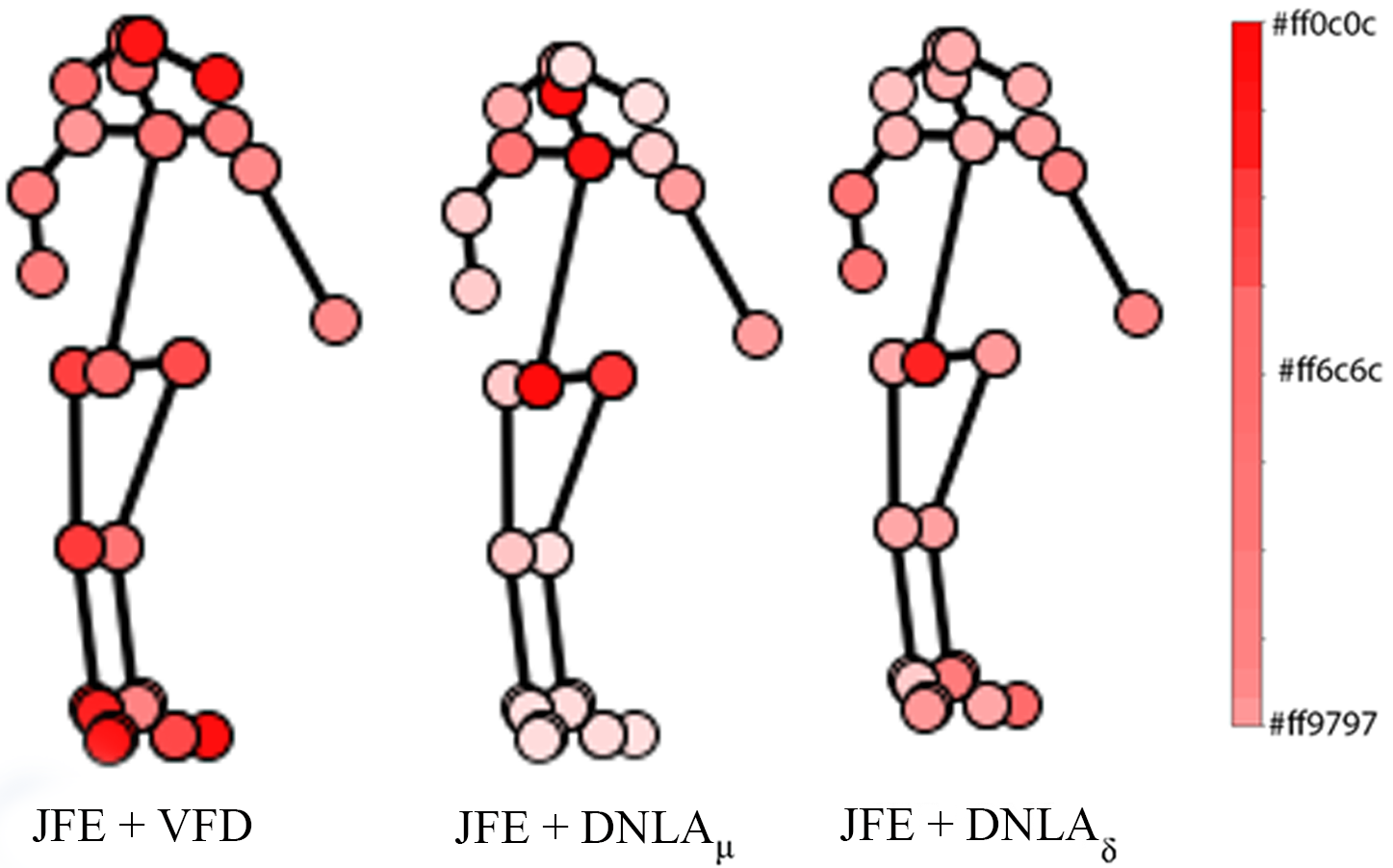}}
\caption{Skeleton heatmap from the last conv layer of different variants of the skeleton model, (a) VFD (b) DNLA$_\mu$(emb) and (c) DNLA$_\delta$(emb). It is evident that compared to VFD which uses similar weights to most of the joints, variants of DNLA can progressively localize joint focus. The DNLA$_\delta$(emb) attends more to limbs and the waist joints, which is reasonable considering gymnastic movements involve these body parts the most.}  
\label{skel_heat}
\end{figure}

Figure~\ref{skel_heat} illustrates the skeleton heatmap from  different variants of the skeleton model. With DNLA, the AQA model was able to focus on the most prominent joints.

Table \ref{AFE_ablate_results} represents experimental results on the RGB videos of AGF-Olympics. AFE + VFD$_\delta$ + MLP achieved the best result. We evaluated two different loss functions, MSE + weighted MAE and USDL from \cite{Tang_2020_CVPR}.

\begin{table}[!htbp]
\scriptsize 
\caption{Results for different architectural choices in appearance stream. The base appearance feature encoder is X3D or I3D. VFD$_{\mu/\delta}$ refers to different kernel sizes to extract discriminative features\label{AFE_ablate_results}.}
\setlength{\tabcolsep}{3pt}
\centering
\begin{tabular}{|p{86pt}|P{35pt}|P{33pt}|P{30pt}|P{35pt}|}
\hline
\textbf{Method} & \textbf{Base Encoder}	& \textbf{SP Corr}	& \textbf{Test loss} & \textbf{Loss fn}\\
\hline
AFE + VFD$_\mu$ + MLP & X3D & 0.57 & 1.05 & L1 + w*L2\\
 \MyIndent + VFD$_\delta$ + MLP & - & 0.68 & 0.57 & L1 + w*L2\\
\hline
 \MyIndent + VFD$_\mu$ + MLP & I3D & 0.61 & 1.89 & USDL\\
\MyIndent + DNLA$_{\mu{(emb)}}$ + MLP & - & 0.64 & 1.24 & USDL\\
\hline
\end{tabular}
\end{table}

\begin{figure}[!tbhp]
\centerline{\includegraphics[width=\columnwidth]{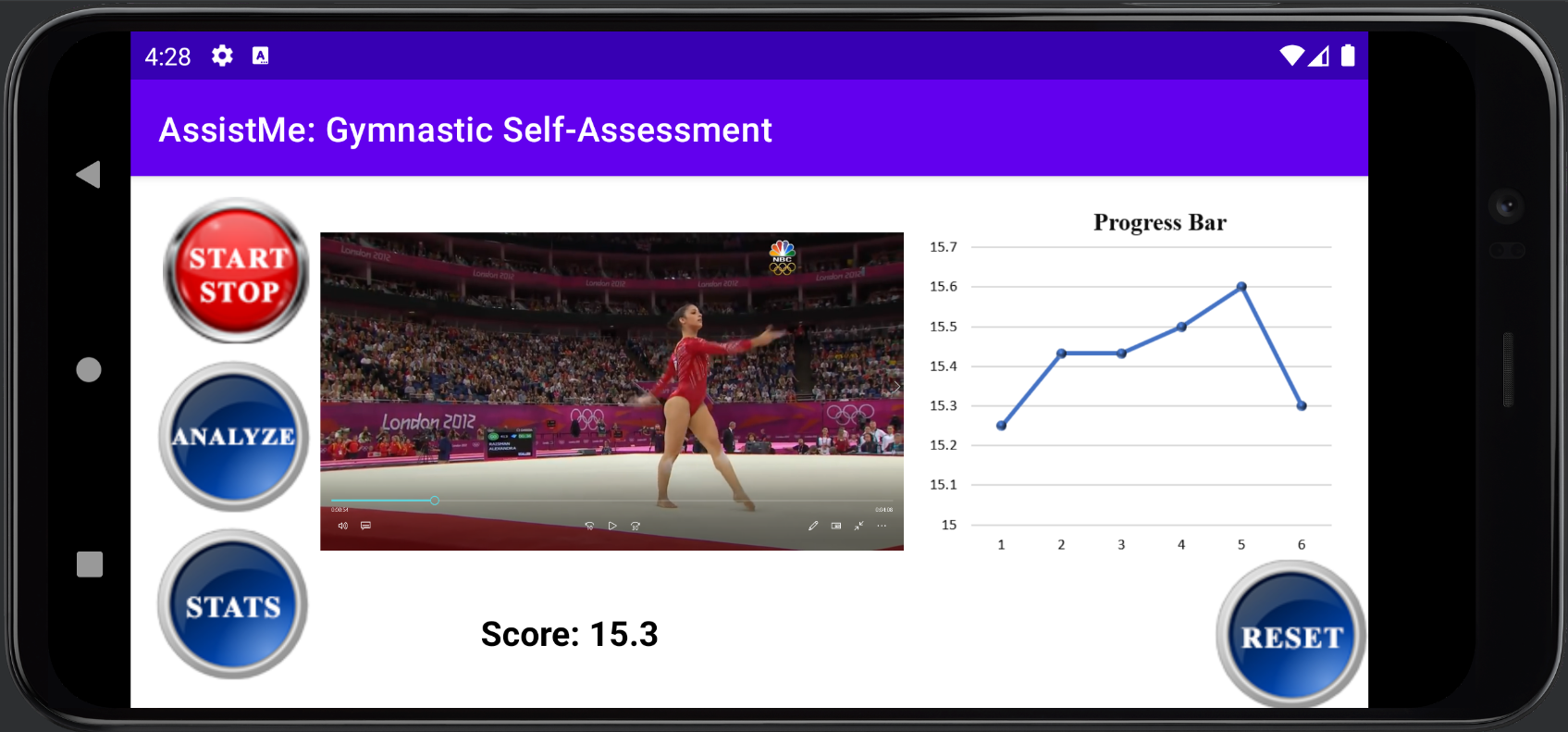}}
\caption{App layout showing a sample video captured and its predicted score. The graph shows score records.} 
\label{app_sample}
\end{figure}

Figure~\ref{app_sample} shows a sample phone app that can be used to communicate with the model deployed on the cloud to assess athletes. The start/stop button starts recording. Pressing the analyze button generates a RestAPI call to the cloud to predict the score. The stats button displays the available score progress chart, and the reset button clears the viewing area.

\section{Conclusion}
We proposed the AGF-Olympics dataset that provides a challenging scenario for long-range temporal modeling in action quality assessment. The complexity of the gymnastic floor combined with the highly random environment in broadcast videos offers a unique situation. Our dataset is substantially more diverse and has subtle intricacies compared to existing action quality assessment datasets. Furthermore, our proposed discriminative non-local attention module presents a new mapping technique to extract extended temporal affinities. Thus it enhances correspondence between isolated activities and overall patterns. With rigorous experiments, we showed that this module significantly improves performance in AQA methods.

\section*{Acknowledgments}
The research is carried out while Sania Zahan is the recipient of a University Postgraduate Award and a University of Western Australia International Fee Scholarship. Professor Ajmal Mian is the recipient of an Australian Research Council Future Fellowship Award (project number FT210100268) funded by the Australian Government.

\bibliographystyle{IEEEtran}
\bibliography{paper}

\newpage
\vspace{11pt}

\begin{IEEEbiography}[{\includegraphics[width=1in,height=1.25in,clip,keepaspectratio]{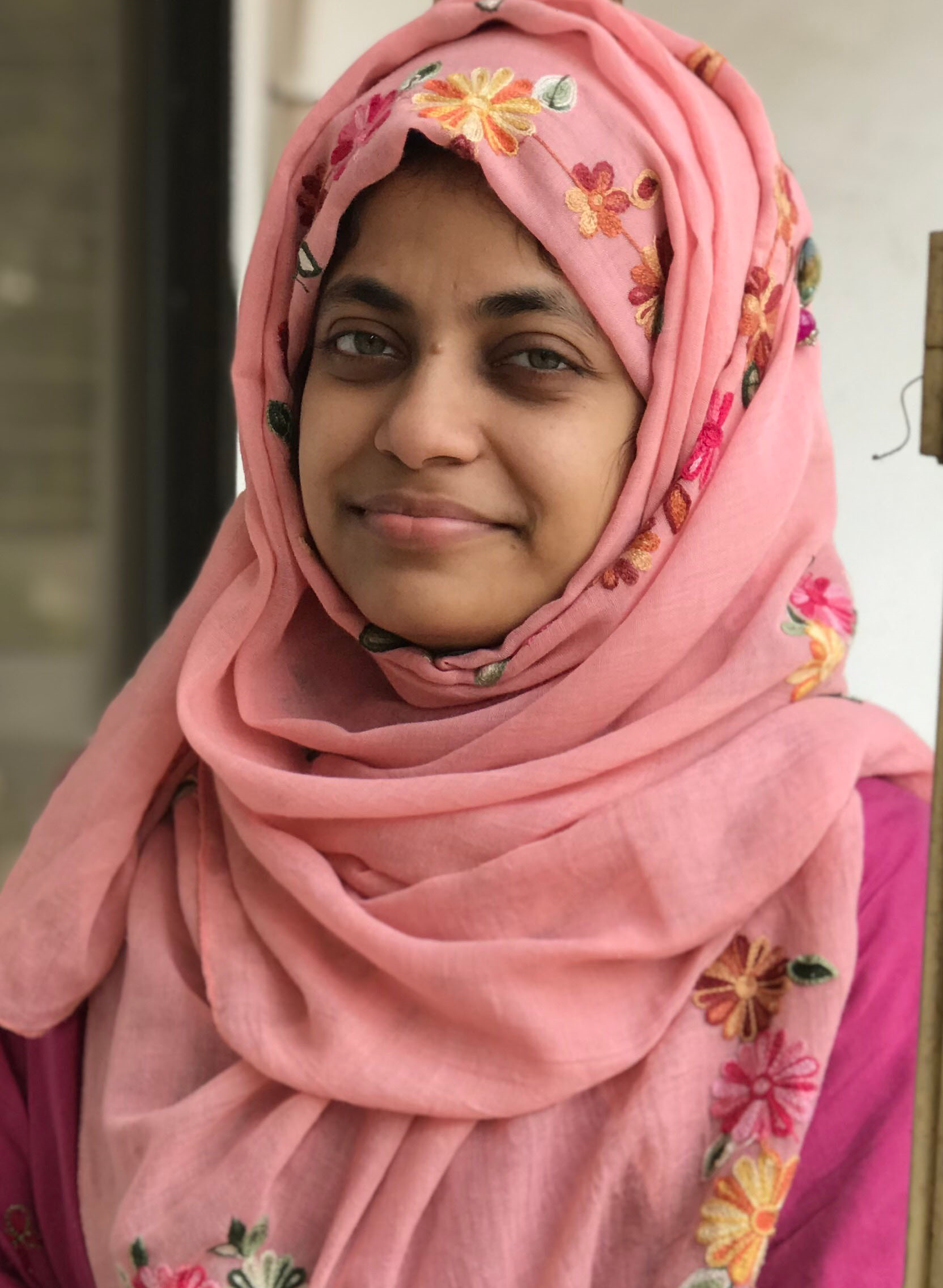}}]{Sania Zahan} is pursuing her Ph.D. degree in Computer Science at the department of Computer Science and Software Engineering at the University of Western Australia (UWA). Sania is the recipient of multiple scholarships and the Varendra University Center for Interdisciplinary Research (CIR) grant 2019-20. Her research interests include computer vision, human action analysis, biomedical signal processing, and deep learning. 
\end{IEEEbiography}

\vspace{11pt}

\begin{IEEEbiography}[{\includegraphics[width=1in,height=1.25in,clip,keepaspectratio]{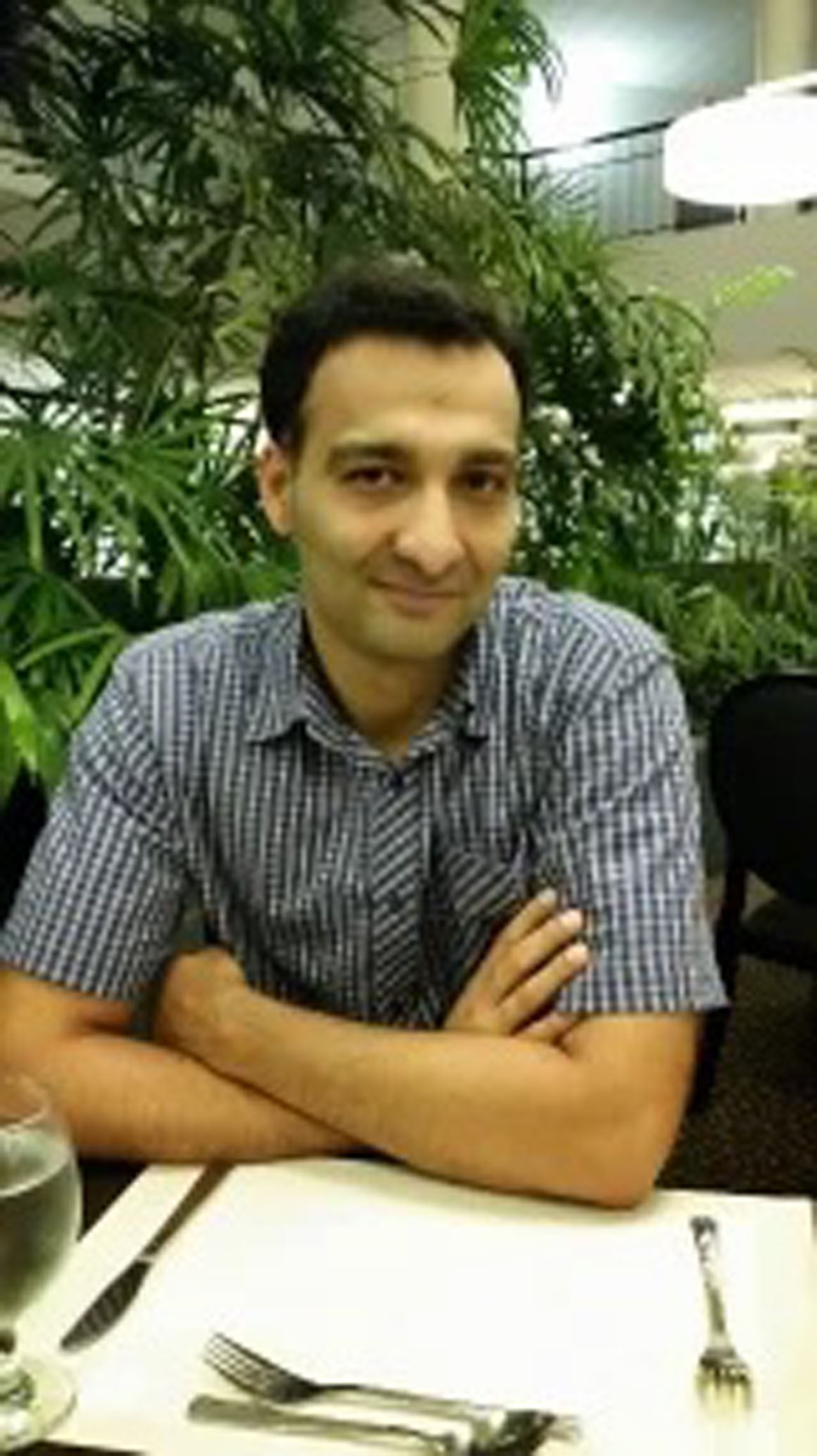}}]{Ghulum Mubashar Hassan} is a faculty member in the Department of Computer Science and Software Engineering at The University of Western Australia. Dr Hassan received his PhD in Computer Science \& Software Engineering and Civil \& Resource Engineering from UWA in 2016. He received his MS and BS from Oklahoma State University, USA, and University of Engineering and Technology, Peshawar, Pakistan, respectively. His research interests are artificial intelligence, machine learning, and their applications in multidisciplinary problems. Dr Hassan is the recipient of multiple teaching excellence and research awards.
\end{IEEEbiography}

\vspace{11pt}

\begin{IEEEbiography}[{\includegraphics[width=1in,height=1.25in,clip,keepaspectratio]{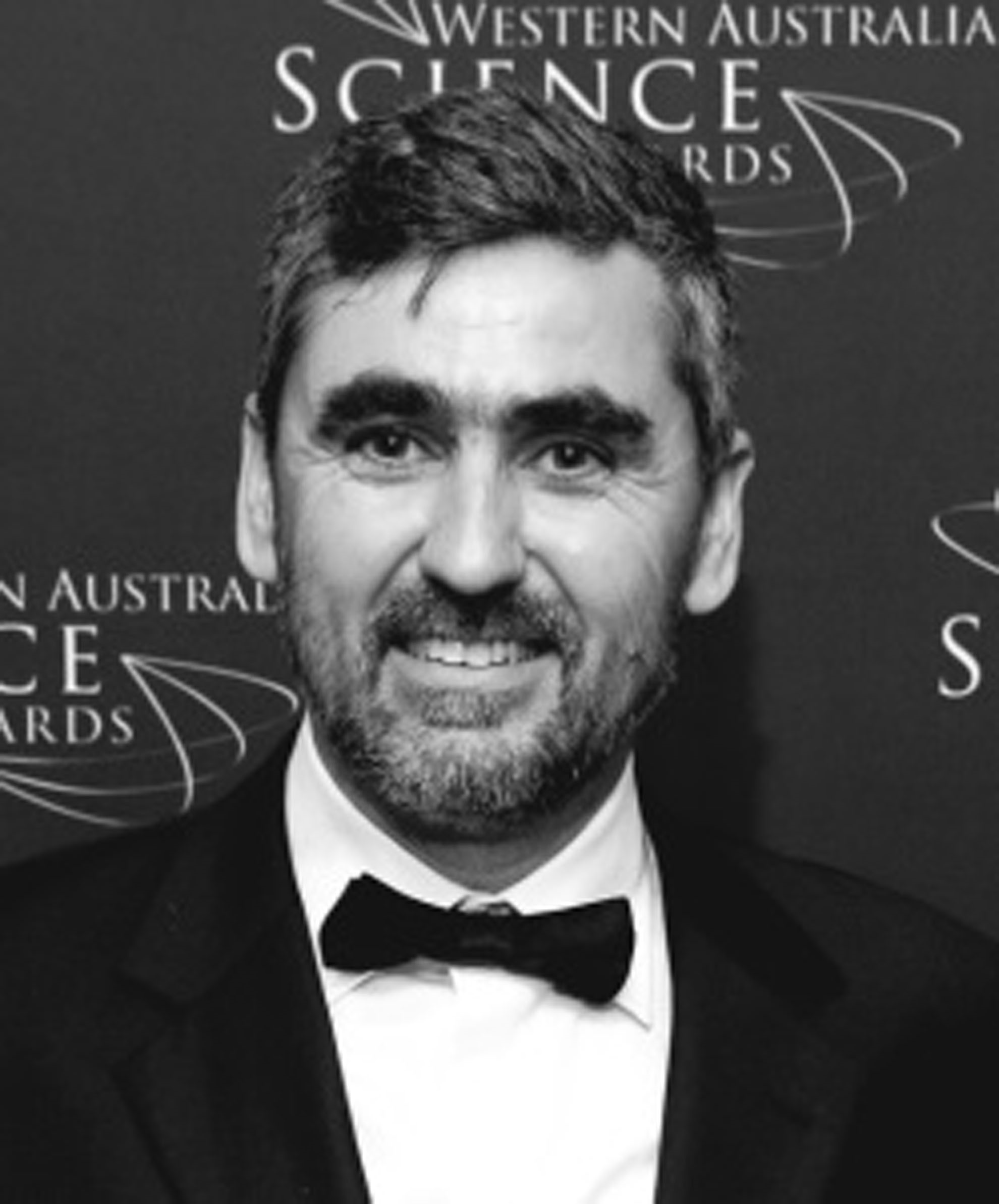}}]{Ajmal Mian} received PhD in Computer Science from UWA in 2007. Ajmal is a Professor of Computer Science at The University of Western Australia. He is the recipient of three prestigious national-level fellowships from the Australian Research Council (ARC) including the Future Fellowship award. He is also a Fellow of the International Association for Pattern Recognition. He received the West Australian Early Career Scientist of the Year Award 2012, the HBF Mid-Career Scientist of the Year Award 2022, and several other awards including the Excellence in Research Supervision Award, EH Thompson Award, ASPIRE Professional Development Award, Vice-chancellors Mid-career Research Award, Outstanding Young Investigator Award, and the Australasian Distinguished Doctoral Dissertation Award. Ajmal Mian has secured research funding from the ARC, the National Health and Medical Research Council of Australia, the US Department of Defence DARPA, and the Australian Department of Defence. He is a Senior Editor for IEEE Transactions on Neural Networks \& Learning Systems and an Associate Editor for IEEE Transactions on Image Processing and the Pattern Recognition journal. He served as a General Chair of the International Conference on Digital Image Computing Techniques \& Applications (DICTA 2019) and the Asian Conference on Computer Vision (ACCV 2018). His research areas include computer vision, deep learning, 3D point cloud analysis, facial recognition, human action recognition, and video analysis.
\end{IEEEbiography}

\vfill

\end{document}